\documentclass{article}

\usepackage{microtype}
\usepackage{graphicx}
\usepackage{subfigure}
\usepackage{booktabs} 
\usepackage{hyperref}

\usepackage[accepted]{icml2023}
\usepackage{amsmath}
\usepackage{amssymb}
\usepackage{mathtools}
\usepackage{amsthm}
\usepackage{amsfonts,mathrsfs,bbm}

\usepackage[capitalize,noabbrev]{cleveref}

\theoremstyle{plain}
\newtheorem{theorem}{Theorem}
\newtheorem{lemma}{Lemma}
\newtheorem{assumption}{Assumption}
\theoremstyle{remark}
\newtheorem{remark}{Remark}

\newcommand{\state}{\mathcal{S}}
\newcommand{\sbr}[1]{\left[#1\right]}

\newcommand{\M}{\mathcal{M}}

\newcommand{\E}{{\mathbb E}}

\newcommand{\A}{\mathcal{A}}

\newcommand{\BR}{\mathfrak{BR}}

\usepackage[disable,textsize=tiny]{todonotes}

\icmltitlerunning{Safe Posterior Sampling for Constrained MDPs with Bounded Constraint Violation}

\begin{document}
\twocolumn[
\icmltitle{Safe Posterior Sampling for Constrained MDPs with Bounded Constraint Violation}
\begin{center}
    \textbf{Krishna C Kalagarla, Rahul Jain, Pierluigi Nuzzo}

Ming Hsieh Department of Electrical and Computer Engineering, University of Southern California, Los Angeles

    Email: {kalagarl,rahul.jain,nuzzo}@usc.edu
\end{center}

\icmlkeywords{Machine Learning, ICML}

\vskip 0.3in
]

\begin{abstract}
Constrained Markov decision processes (CMDPs) model  scenarios of sequential decision making with multiple objectives that are increasingly important in many applications. However, the model is often unknown and must be learned online while still ensuring the constraint is met, or at least  the violation is bounded with time. Some recent papers have made progress on this very challenging problem but either need unsatisfactory assumptions such as knowledge of a safe policy, or have high cumulative regret. We propose the Safe PSRL (posterior sampling-based RL) algorithm that does not need such assumptions and yet performs very well, both in terms of theoretical regret  bounds  as well as empirically. The algorithm achieves an efficient tradeoff between exploration and exploitation by use of the posterior sampling principle, and provably suffers only bounded constraint violation by leveraging the idea of pessimism. Our approach is based on a primal-dual approach.  We establish a sub-linear $\tilde{\mathcal{
O}}\left(H^{2.5} \sqrt{|\state|^2 |\A| K} \right)$ upper bound on the Bayesian reward objective regret along with a \emph{bounded}, i.e., $\tilde{\mathcal{O}}\left(1\right)$ constraint violation regret over $K$ episodes for an $|\state|$-state, $|\A|$-action, and horizon $H$ CMDP. 
\end{abstract}

\section{Introduction}
\label{sec:intro}

Markov decision processes (MDPs)~\cite{Puterman:1994:MDP:528623} are used to model many scenarios involving sequential decision-making. They are used in a wide variety of settings like robotics, cyber-physical systems, and safety-critical autonomous vehicles. However, a traditional reinforcement learning (RL) formulation which seeks to maximize a single cumulative reward cannot capture many problems that have multiple objectives. For example, this is the case for a robot that needs to perform a certain reward-yielding task while ensuring that the average energy expended is bounded below a threshold. Such scenarios are well-modeled by constrained MDPs (CMDPs) ~\cite{altman1999constrained}, which extend the MDP formalism by considering additional constraints on the expected cumulative performance of a policy. In the CMDP setting, one seeks to find an optimal policy which maximizes the  cumulative objective reward while satisfying constraints on cost objectives.

In this paper, we consider the problem of online learning for finite-horizon CMDPs, where an agent interacts with the environment repeatedly in episodes of fixed length. The transition probability is not known to the agent, thereby requiring the agent to learn about the system dynamics by observing the  past states and actions. The performance of this agent is measured by the notion of \emph{cumulative regret}, i.e., the difference between the cumulative reward of the learning agent and that of the optimal policy. 

This online learning problem thus leads to the well-known trade-off between \emph{exploration} and \emph{exploitation}: should the agent \emph{explore} the environment to improve future performance, or \emph{exploit} the current knowledge for better short-term performance? 

A common approach to balance this exploration-exploitation trade-off is the \textit{`Optimism in the Face of Uncertainty'} (OFU) principle \cite{lai1985asymptotically}. The idea is that each state and action are assigned \textit{optimism bonuses} based on the current knowledge. The agent then chooses a policy which maximizes the expected return under this \textit{optimistic} model.  The bonuses are designed to promote exploration of poorly-understood state-action pairs. This approach has been widely used for online learning in MDPs \cite{jaksch2010near,azar2017minimax,jin2018q,wei2020model,kalagarla2021sample}. 

Another alternative for efficient exploration is \textit{posterior sampling} (also called Thompson sampling) \cite{thompson1933likelihood}. In this approach, a posterior distribution is maintained over the unknown transition probability model based on the prior distribution and dataset about visited trajectories. At the beginning of each episode, a model is sampled from this posterior distribution. The agent then chooses a policy which is optimal with respect to the sampled model and follows it for the duration of the episode. The advantages of posterior sampling over OFU stem from the fact that (i) known information about the model can be incorporated into the algorithm through the prior distribution, and (ii) posterior sampling algorithms have demonstrated superior empirical performance for online learning over OFU-type algorithms including in the RL setting \cite{osband2013more,ouyang2017learning}.

In this paper, we use the posterior sampling approach and introduce the \texttt{Safe PSRL} algorithm for efficient exploration in the finite-horizon CMDP setting. Our algorithm  uses the primal-dual approach for CMDPs wherein the primal part performs unconstrained MDP planning with a sampled transition probability, and the dual part updates the Lagrangian variable to track the constraint violation.

We achieve bounded constraint violation regret by leveraging the idea of \emph{pessimism}, introduced earlier in the context of constrained bandits \cite{liu2021efficient}. ``Pessimism" is achieved by tightening the constraint of the CMDP problem in every episode at decreasing levels. The key is, however, how this tightening is achieved. By appropriately balancing exploration via posterior sampling and \emph{safe} learning via pessimism, we show that the \texttt{Safe PSRL} algorithm achieves sub-linear $\tilde{\mathcal{
O}}\left(\frac{H^{2.5}}{\tau - c^0} \sqrt{|\state|^2 |\A| K} \right)$ reward regret while achieving bounded, i.e., $\tilde{\mathcal{
O}}(1)$-constraint violation regret for an $|\state|$-state, $|\A|$-action, and an $H$ episode length CMDP over $K$ number of episodes.  $\tau$ denotes the desired threshold for constraint violation, and $c^0$ is a known feasible expected cumulative constraint cost of the CMDP.

The contributions of this paper are the following: (i) We present the first PSRL algorithm for CMDPs that not only achieves  $\tilde{\mathcal{
O}}\left(\sqrt{K}\right)$ objective reward regret but also $\tilde{\mathcal{
O}}(1)$ constraint violation regret.  Unlike other proposed algorithms, our algorithm does not need knowledge of a constraint-satisfying \textit{safe} policy. (ii) Our \texttt{Safe PSRL} algorithm has better empirical performance than state-of-the-art OFU-type algorithms for the same setting introduced in \cite{liu2021learning,bura2022dope} which need  knowledge of a safe policy. (iii) The algorithm design is simpler than other state-of-the-art algorithms \cite{liu2021learning,bura2022dope} for the problem: The key design choice are two pessimism parameters. The regret analysis involves a novel decomposition which allows us to leverage posterior sampling regret analysis and Lyapunov-drift analysis for the dual variables.

\section{Related Work}
 
Posterior (or Thompson) sampling goes back to the work of \cite{thompson1933likelihood}, but attracted less attention for several decades until empirical evidence \cite{chapelle2011empirical} showed its superior performance for online learning. Recently, it has been widely applied to various settings like multi-armed bandits \cite{kaufmann2012thompson,agrawal2012analysis,agrawal2013thompson}, MDPs \cite{osband2013more,gopalan2015thompson,osband2017posterior,ouyang2017learning} and POMDPs \cite{jafarnia2021online}.

Multi-armed bandits (MAB) are a special case of MDPs with a single state and unknown reward function. Safe online learning has been studied for MABs in multiple settings~\cite{amani2019linear,khezeli2020safe,pacchiano2021stochastic,liu2021efficient}. However,  in comparison, the multi-step and multi-state MDP settings are more challenging due to the unknown transition probability. 

In the CMDP setting, several existing works~\cite{efroni2020exploration,qiu2020upper,agarwal2022regret} leverage OFU or posterior sampling to provide $\tilde{\mathcal{O}}(\sqrt{K})$ regret for the reward as well as the constraint objective, where $K$ is the number of episodes. Such an approach, however, can lead to a large number of constraint violations during learning, which is unacceptable during various safety-critical tasks such as driving or power distribution. Thus, the problem of an online RL algorithm that has sublinear (and hopefully, $ \tilde{\mathcal{
O}}(\sqrt{K})$) reward regret while achieving \textit{bounded} constraint violation regret,   particularly one that also has very good empirical performance, remains open.

OFU-based algorithms have been widely used for efficient learning in CMDPs, e.g., in the setting of PAC performance guarantees for finite-horizon CMDPs~ \cite{hasanzadezonuzy2021learning,kalagarla2021sample}, or to provide regret bounds for CMDPs in the finite-horizon setting \cite{efroni2020exploration,brantley2020constrained} and infinite-horizon average cost setting \cite{singh2020learning}. Policy gradient algorithms for CMDPs \cite{ding2020natural,ding2021provably} have also been studied. However, these algorithms do not provide bounded or zero constraint violation guarantees.

Recently, some OFU-based approaches for \emph{safe} learning with bounded or zero constraint violation guarantees have been proposed \cite{zheng2020constrained, chen2022learning,bai2022achieving,wei2022triple,liu2021learning}. But these either assume the transition model is known (but reward function is not), or only satisfy the constraint with high probability, or assume that a safe policy is known to the algorithm (and can be used by it), e.g., in \cite{liu2021learning,bura2022dope}. The OptPess-PrimalDual algorithm in \cite{liu2021learning} is the closest comparable algorithm but our \texttt{Safe PSRL} algorithm is better in terms of its dependence on various problem parameters, e.g., it has $\tilde{\mathcal{
O}}(H^3|\state|^{1.5})$ dependence as opposed to $\tilde{\mathcal{
O}}(H^{2.5}|\state|)$ for ours.
 
While the use of the posterior sampling principle for constrained RL  problems is under-explored (despite the promise of better empirical performance), \cite{agarwal2022regret} indeed introduces a PSRL algorithm for CMDPs but for the average setting. Moreover, it only achieves a  $\tilde{\mathcal{O}}(\sqrt{K})$ constraint violation regret which is worse than our $\tilde{\mathcal{O}}(1)$ bound.

\section{Preliminaries}
\label{sec:prelims}
\subsection{Notation} 

We denote the probability simplex over set $S$ by $\Delta_{S}$. We use the notation $\tilde{\mathcal{O}} $ which has similar meaning as the usual $\mathcal{O}$ notation but ignores logarithmic factors.

\subsection{Finite-Horizon MDPs} An episodic finite-horizon MDP~\cite{Puterman:1994:MDP:528623} can be formally defined by a tuple $\M = (\state,\A,H,s_{1},p,r)$, where $\state$ and $\A$ denote the state and action spaces, respectively. In this setting, the agent interacts with the environment in episodes of fixed length $H$, with each episode starting with a random  initial state denoted $s_{1}$. The non-stationary transition probability $p_{h}(s'|s,a)$  is the the probability of transitioning to state $s'$ on taking action $a$ at state $s$ at time step $h\in \left[ 1: H \right]$ of the episode. The non-stationary reward obtained on taking action $a$ in state $s$ at time step $h$ of an episode is denoted by a random variable $R_{h}(s,a) \in \left[ 0, 1 \right] $, with mean $r_{h}(s,a)$. We use $r $ as a shorthand to denote the  mean reward vector $r_1,\ldots,r_H $.

A non-stationary randomized policy $\pi = (\pi_{1}, \ldots , \pi_{H}) \in \Pi$ where $\pi_{i} : \state \to \Delta_{\A}$, maps each state to a probability simplex over the action space $\A$. The action $a_{h}$ at time step $h$ at state $s_h$ is taken according to the policy $\pi$, $a_{h} \sim \pi_{h}(s_h)$. The value function of a non-stationary randomized policy $\pi$, $V_{h}^{\pi}(s;r,p)$  (when clear, $s, r,$ and $p$ are omitted) at a state $s \in \state$ and time step $h \in \left[ 1 : H \right]$ is defined as:

$$V_{h}^{\pi}(s;r,p) := \E_{\pi} \left[\sum_{i=h}^{H} r_{i}(s_{i},a_{i}) | s_h = s, p \right], $$
where the expectation is over the distribution induced by the environment and policy randomness. Similarly, the Q-value function of a policy $\pi$, $Q_{h}^{\pi}(s,a;r,p)$, for a state $s \in \state$, an action $a \in \A$ and time step $h \in \left[ 1 : H \right]$, is defined as 
\begin{align}
   &Q_{h}^{\pi}(s,a;r,p) :=
 r_{h}(s,a) + \\ 
 &\E_{\pi} \left[\sum_{i=h+1}^{H} r_{i}(s_{i},a_{i}) | s_h = s,a_h = a, p \right]. \notag
\end{align}

We can always find an optimal non-stationary deterministic policy $\tilde{\pi}$ \cite{Puterman:1994:MDP:528623} such that $V_{h}^{\tilde{\pi}}(s) = \tilde{V}_{h}(s) = \text{sup}_{\pi} V_{h}^{\pi}(s)$ and 
$Q^{\tilde{\pi}}_{h}(s,a)  =\tilde{Q}_{h}(s,a)= \text{sup}_{\pi} Q_{h}^{\pi}(s,a)$.
The optimal policy can be computed by using 
backward induction on the Bellman optimality equations \cite{Puterman:1994:MDP:528623}:
\begin{equation}\label{eq:backind}
   \begin{aligned}
      & \tilde{V_{h}}(s) = \text{max}_{a \in \A} \left[r_{h}(s,a) + p_{h}(\cdot|s,a)\tilde{V}_{h+1}\right],\\
      &\tilde{Q}_{h}(s,a) =  r_{h}(s,a) + p_{h}(\cdot|s,a)\tilde{V}_{h+1},
 \end{aligned}  
\end{equation}

 where $\tilde{V}_{H+1}(s) = 0$ and $\tilde{V}_{h}(s) = \text{max}_{a \in \A}\tilde{Q}_{h}(s,a)$. The optimal policy $\tilde{\pi}$ is then greedy with respect to $\tilde{Q}_{h}$.

 \subsection{Finite-Horizon Constrained MDPs}
 
 A finite-horizon constrained MDP (CMDP) \cite{altman1999constrained} is a finite-horizon MDP  with a required upper bound on expectation of  on  a cost function, $\{c,\tau \in (0,H ] \}$. The non-stationary cost obtained on taking action $a$ in state $s$ at time step $h \in \left[ 1: H \right]$ with respect to the constraint cost function is denoted by a random variable $C_{h}(s,a) \in \left[ 0, 1 \right] $, with mean $c_{h}(s,a)$. Similar to $r$, we use $c$ to denote the mean cost  vector $  c_1,\ldots,c_H $. 
 
 The total expected reward (cost) of an episode under policy $\pi$ with respect to the reward (cost) function $r$ ($c$) is the respective value function from the initial state $s_1$,  i.e., $V_{1}^{\pi}(s_1;r,p) (V_{1}^{\pi}(s_1;c,p)) $ (by definition). Our objective in this CMDP setting is to find a policy which maximizes the total expected objective reward under the constraint that the total expected constraint cost is below a desired threshold. Formally,
\begin{equation}\label{eq:obj}
    \begin{aligned}
   \pi^{*} \in \underset{\pi \in \Pi }{\text{ argmax }} \quad & V_{1}^{\pi}(s_1;r,p)\\ \textrm{s.t.} \quad  & V_{1}^{\pi}(c,p) \leq \tau.
\end{aligned}
\end{equation}
The optimal value is denoted by $V^{*}(s_1; r, p) = V_{1}^{\pi^{*}}(s_{1};r,p) $. A deterministic optimal policy may not exist, hence we need to consider $\Pi$ to be the class of all randomized policies \cite{altman1999constrained}. Since the Bellman optimality equations may not hold due to the constrained nature of the problem, we cannot leverage dynamic programming-based backward induction algorithms to find an optimal policy. However, a linear programming approach can be given that will find an optimal policy  \cite{altman1999constrained}.

\section{The Learning Problem}\label{sec:problem}

We consider the setting where an agent repeatedly interacts with a finite-horizon CMDP $\M = (\state,\A,H,s_1,p,r,\{c,\tau \})$ over multiple episodes of fixed length $H$, starting each episode from the same initial state $s_1$ and with stationary transition probability (i.e., $p_h = p, \forall h$). We employ the Bayesian framework and regard the transition probability $p$ as random with a prior distribution $\mu_1$. The realized transition probability is unknown to the learning agent. We consider finite-horizon CMDP whose transition probability lies in the set $\Theta_{c_0}$ with the following property:

\begin{assumption} \label{ass: class of cmdp}
For all $\hat{p} \in \Theta_{c_0}$, there exists a policy $\pi_{0}^{\hat{p}}$ such that $V_{1}^{\pi_{0}^{\hat{p}}}(c,\hat{p}) \leq c_0 < \tau$.
\end{assumption}

Moreover, we assume that the support of the prior distribution $\mu_1$ is a subset of $\Theta_{c_0}$ and $c_0$ is known. Without loss of generality,\footnote{The complexity of learning the cost and reward functions is dominated by the complexity of learning the transition probability \cite{auer2005online}. The algorithm can be readily extended to the setting of unknown cost and reward functions by using their empirical estimate in place of the known cost and reward functions.}  we assume that the reward and cost functions $r$ and $c $ are respectively are known to the learning agent. Note that the above assumption is not only reasonable but also necessary to ensure the problem is feasible.

The agent interacts with the environment for $K$ episodes, each of length $H$. In each episode, the agent starts from a state $s_1$ and chooses a Markov policy $\pi_k$ determined by the information gathered until that episode. This policy is then executed until the end of the episode, while collecting the rewards and costs. The main objectives of the learning agent are to:
 
 (1) Maximize the expected cumulative reward after K episodes or equivalently, minimize the Bayesian regret with respect to the reward function defined as:
\begin{align}
    \BR(K;r) := \E\sbr{\sum_{k = 1}^K \left( V_{1}^{\pi^*}(s_1; r, p) - V^{\pi^k}_{1}(s_1; r, p) \right)}.
    \label{eq:regret_rew}
\end{align}

(2) Minimize the constraint violation or equivalently, minimize the Bayesian regret with respect to the constraint defined as:
\begin{align}
    \BR(K;c) := \E\sbr{\sum_{k = 1}^K \left(  V^{\pi^k}_{1}(s_1; c, p) - \tau \right)}.
    \label{eqn:regret_cost}
\end{align}

With respect to these objectives, we propose an algorithm which is able to achieve sub-linear regret with respect to the reward objective while ensuring that regret with respect to the cost constraint is bounded above by a constant, i.e., independent of the number of episodes $K$.

\section{The Safe PSRL Algorithm}
\label{sec:algo}

We propose the Safe Posterior Sampling-based Reinforcement Learning (\texttt{Safe PSRL}) algorithm for the finite-horizon CMDP model. This algorithm leverages the idea of posterior sampling to balance exploration and exploitation. It also takes a primal-dual approach to handle the constraint cost objective along with reward maximization objective. 

We further introduce the idea of pessimism \cite{liu2021efficient} to ensure that the cost regret is bounded. This ``pessimism" is achieved by considering a ``more constrained" CMDP problem as compared to the original problem. This is done by decreasing the threshold by $\epsilon_k$ in each episode $k$. Formally,
we consider the objective:
\begin{equation}\label{eq:tobj}
    \begin{aligned}
   {\text{ max }} \quad & V_{1}^{\pi}(r,p)\\ \textrm{s.t.} \quad  & V_{1}^{\pi}(c,p) \leq \tau - \epsilon_k.
\end{aligned}
\end{equation}

This pessimistic term $\epsilon_k$ ensures bounded cost regret and it decreases as the episode count increases. 

The algorithm starts with the prior distribution $\mu_1$ on the transition probability. Then, at every time step $t$, the learning agent maintains a posterior distribution $\mu_t$ on the unknown transition probability $p$ given by $\mu_t(\Theta) = \mathbb{P}(p \in \Theta | \mathcal{F}_t)$ for any set $\Theta \subseteq \Theta_{c_0}$. Here $\mathcal{F}_t$ is the information available at time $t$, i.e., the sigma algebra generated by encountered states and actions upto time $t$, $(s_1, a_1, \cdots, s_{t-1}, a_{t-1}, s_t)$. On observing the next state $s_{t+1}$ by taking action $a_t$ at state $s_t$, the posterior is updated according to Bayes's rule: 
\begin{align}
\label{eq: update rule}
\mu_{t+1}(dp) = \frac{p_{t}(s_{t+1}|s_t, a_t)\mu_t(dp)}{\int p^{'}_{t}(s_{t+1}|s_t, a_t)\mu_t(dp')}.
\end{align}

In parallel, the algorithm proceeds as follows: At the beginning of each episode $k$, transition probability $\hat{p}_{k}$ is sampled from the posterior distribution $\mu_{t_k}$ (where $t_k$ is the time step corresponding to beginning of episode $k$).  We then consider the Lagrangian defined as:
\begin{align*}
    L_k(\pi, \lambda) := V^{\pi}_1(r,\hat{p}_{k}) + \frac{\lambda_k}{\eta_k} \left( \tau - \epsilon_k - V_1^{\pi}(c, \hat{p}_{k}) \right),
\end{align*}
The learning agent then chooses a Markov policy $\pi_k$ (primal update) which maximizes the above Lagrangian. We can find such a policy by applying standard dynamic programming with respect to the reward function $r - \frac{\lambda_k}{\eta_k}c $. The (dual) parameter $\lambda_k$ is updated according to the sub-gradient algorithm as follows:
\begin{align*}
    \lambda_{k+1} = \left(\lambda_k + V_1^{\pi_k}(c, \hat{p}_k) + \epsilon_{k} - \tau\right)_{+}
\end{align*}
The agent then applies the policy $\pi_k$ for the $H$ steps of episode $k$. 

We note that while some of the details of the algorithm are natural (as they are common to PSRL algorithms for various settings) \cite{ouyang2017learning,jafarnia2021online, jafarni2021online,jafarnia2021learning}, the key novelty in the design are the $\epsilon_k$  and $\eta_k$ parameters to be used in conjunction with a primal-dual approach. Their choice is guided by the regret analysis presented in Section \ref{sec:analysis}.  

The \texttt{Safe PSRL} algorithm is summarized next. 

\begin{algorithm}[H]
\label{alg:Safe PSRL}
\caption{{\tt Safe-PSRL}}
\begin{algorithmic}
\STATE \textbf{Input}: $K, \mu_{1}, c_0,\tau$
\STATE \textbf{Initialization}: $ \lambda^{1} \leftarrow 0 $
\FOR{episodes $k=1,\ldots,K$}
\STATE $\epsilon_k \leftarrow \frac{5|H|^{1.5}\sqrt{|\state|^{2}|\A|}(\log k|\state||\A|H + 1)}{\sqrt{k \log k |\state||\A|H}}$
\STATE $\eta_k \leftarrow (\tau - c_0) H \sqrt{k}$\\
\STATE $t_k = (k-1)H + 1$
\STATE Generate $\hat{p}_k \sim \mu_{t_k}(.)$ \\
\STATE
\STATE  Compute $\pi_{k} \in \arg \max_{\pi} V^{\pi}_{1}(r - \frac{\lambda_k }{\eta_k } c,\hat{p}_k)$ according to \eqref{eq:backind} (Policy Update) \\
\STATE  $\lambda_{k+1} \leftarrow \max(0, \lambda_k + V^{\pi_k}_{1}(c,\hat{p}_k) + \epsilon_k - \tau )$ (Dual Update)\\
\STATE
\FOR{$t = (k-1)H + 1,\ldots,kH$}
\STATE Choose action $a_t \sim  \pi_{k}(s_t)$ \\
\STATE Observe $s_{t+1} \sim p(.|s_t,a_t)$
\STATE Update the posterior distribution $\mu_{t+1}$ according to \eqref{eq: update rule}
\ENDFOR
\ENDFOR
\end{algorithmic}
\end{algorithm}
The following theorem then establishes  that the \texttt{Safe PSRL} algorithm can achieve sub-linear $\tilde{\mathcal{
O}}(\sqrt{K})$ reward regret while achieving bounded constraint violation regret.

\begin{theorem}\label{thm:regret}
Suppose Assumption \ref{ass: class of cmdp} holds, then the reward and cost regret of the \texttt{Safe PSRL} algorithm is upper bounded as:
\begin{align*}
        \BR(K;r) &= \tilde{\mathcal{
O}}\left(\frac{H^{2.5}}{\tau - c^0} \sqrt{|\state|^2 |\A| K} \right),~\text{and}\\
        \BR(K;c) &= \tilde{\mathcal{
O}}\left(C''(H-\tau) + H^{1.5} \sqrt{|\state|^2 |\A| C''} \right) \\
&= \mathcal{O}(1),
    \end{align*}
    where $C'' = \mathcal{O} (\frac{H^3 |\state|^2 |\A|}{(\tau - c^0)^2} )$ is independent of $K$.
\end{theorem}

\begin{remark}
(i) We note that the upper bound on $\BR(K;r)$  of the \texttt{OptPess-PrimalDual} algorithm \cite{liu2021learning} is $\tilde{\mathcal{
O}}\left( H^{3} \sqrt{|\state|^3|\A|K}\right)$. Thus, our upper bound is the same in terms of $|\A|$,$K$ and better in terms of $|\state|$ and $H$. Both, OptPess-PrimalDual and \texttt{Safe PSRL} algorithms achieve $\tilde{\mathcal{
O}}\left(1 \right)$ upper bounds on $\BR(K;c)$. 

(ii) We note that the upper bound on $\BR(K;r)$  of the \texttt{DOPE} algorithm \cite{bura2022dope} is $\tilde{\mathcal{
O}}\left( H^{3} \sqrt{|\state|^2|\A|K}\right)$. Thus, our upper bound is the same in terms of $|\A|,K,|\state|$ and better in terms of $H$. While our bounds are comparable to those of \texttt{DOPE}, we shall see that the numerical performance is much better. Further, the \texttt{DOPE} algorithm guarantees zero constraint violations with high probability. But, this requires a strong assumption, i.e, knowledge of a safe policy that can satisfy the constraint.

(iii) The CMDP-PSRL algorithm \cite{agarwal2022regret} uses posterior sampling in the average CMDP setting and achieves $\tilde{\mathcal{
O}}\left( T_{M} |\state| \sqrt{|A|K}\right)$ reward objective and the same constraint violation regret, where $T_M$ is the mixing time. In comparison, we are able to achieve bounded constraint violation regret. 
\end{remark}

\section{Regret Analysis}
\label{sec:analysis}

We now provide theoretical analysis of the \texttt{Safe PSRL} algorithm by providing details of the proof of Theorem \ref{thm:regret}.
We first state some relevant results from the literature on posterior sampling in the context of reinforcement learning.

A key property of posterior sampling \cite{osband2013more} \\ is the posterior sampling lemma, i.e.,  the transition probability $\hat{p}_t$ sampled from the  posterior distribution at time $t$ and transition probability $p$ have the same distribution.

\begin{lemma}\label{lem: pseq}
For any function $f$, we have $\E \sbr{f(\hat{p}_t)} = \E \sbr{f(p)}$ where $p$ is the transition probability (with the prior distribution $\mu_1$) and $\hat{p}_t$ is the sampled transition probability from the posterior distribution $\mu_t$ at time $t$.
\end{lemma}

The following is a restatement \cite{osband2013more} of the sub-linear regret bound achieved when using posterior sampling for unconstrained finite horizon MDPs.

\begin{lemma}\label{lem: psrlreg}
\cite{osband2013more} The Bayesian regret of the PSRL algorithm for unconstrained MDPs is given by
\begin{equation}
\begin{split}
& \sum_{k = 1}^K \E\sbr{V^{\pi^k}_{1}(c, p) - V^{\pi^k}_{1}(c, \hat{p}_k)}\\ & \leq H^{1.5}\sqrt{30|\state|^2|\A|K\log(|\state||\A|KH)} + 2H.
\end{split}
\end{equation}
\end{lemma}

\subsection{Cost Constraint Violation Analysis}

We first present analysis of the cost constraint violation. We can decompose the constraint violation regret as follows:
\allowdisplaybreaks\begin{align}
\label{eq:cregret}
    &\BR(K;c) := \E\sbr{\sum_{k = 1}^K \left(  V^{\pi^k}_{1}(c, p) - \tau \right)}\notag\\
    &= \sum_{k = 1}^K \E\sbr{V^{\pi^k}_{1}(c, p) - V^{\pi^k}_{1}(c, \hat{p}_k) +V^{\pi^k}_{1}(c, \hat{p}_k) - \tau }\notag\\
    &= \sum_{k = 1}^K \E\sbr{V^{\pi^k}_{1}(c, p) - V^{\pi^k}_{1}(c, \hat{p}_k)} + \sum_{k = 1}^K \E\sbr{V^{\pi^k}_{1}(c, \hat{p}_k) - \tau}\notag\\
    &\leq \sum_{k = 1}^K \E\sbr{V^{\pi^k}_{1}(c, p) - V^{\pi^k}_{1}(c, \hat{p}_k)} 
 + \sum_{k = 1}^K \E\sbr{\lambda_{k+1} - \lambda_k - \epsilon_k}\notag\\ 
 &\text{ (by dual update rule of algorithm)}\notag\\
 &= \sum_{k = 1}^K \E\sbr{V^{\pi^k}_{1}(c, p) - V^{\pi^k}_{1}(c, \hat{p}_k)}  + \E\sbr{\lambda_{K+1}}- \sum_{k = 1}^K \epsilon_k \\
 & \leq H^{1.5}\sqrt{30|\state|^2|\A|K\log(|\state||\A|KH)} + 2H\notag\\
 & + \E\sbr{\lambda_{K+1}} - \sum_{k = 1}^K \epsilon_k \label{eq:firstsum}
 \end{align}
where the last upper bound follows by use of Lemma \ref{lem: psrlreg} to upper bound the first term in \eqref{eq:cregret}.

We next show that the dual parameter $\E\sbr{\lambda_{K+1}}$ can be upper bounded by use of Lyapunov-drift analysis. To that end, we restate the following lemma \cite{liu2021efficient} which states the Lyapunov-drift conditions for the boundedness of a random process.

\begin{lemma}\label{lem:bandbound}
\cite{liu2021efficient} Consider a random process $S(t)$ with a 
Lyapunov function $\Phi(k)$ such that $\Phi(0) = 0$ and $\Delta(k) = \Phi(k+1) - \Phi(k) $ is the Lyapunov drift. Given an increasing sequence $\{ \varphi_k\}$ and constants $\rho $ and $\nu_{max}$ with
$0 < \rho \leq \nu_{max}$, if the expected drift $\E \sbr{\Delta(k)|S(k) = s}$ satisfies the following conditions:\\
(i) There exists constants $\rho>0$ and $\varphi_{k}>0$ s.t. $\E\sbr{\Delta(k)|S(k) = s} \le-\rho$ when $\Phi(k) \ge \varphi_{k}$, and \\
    (ii) $|\Phi(k+1)-\Phi(k)| \le \nu_{\max}$ holds with probability $1$, then
    \begin{align*}
    \E\sbr{ e^{\zeta \Phi(t)}} \le \E\sbr{e^{\zeta \Phi_{0}}}+\frac{2 e^{\zeta\left(\nu_{\max }+\varphi_{t}\right)}}{\zeta \rho},
    \end{align*}
    where $\zeta=\rho/(\nu_{\max }^{2}+\nu_{\max} \rho / 3)$.
\end{lemma}

We divide the episodes into two parts,  i.e. $k <  C''$ and $k \geq C'' $ where $C'' = \frac{80H^3|\state|^2|\A|}{(\tau-c_0)^2}$. We can clearly see that for 
$k \geq C''$, we have $\epsilon_k \leq \frac{\tau - c
_0}{2}$. Thus, for $k \geq C''$, Problem \eqref{eq:tobj} is feasible for all $\hat{p_k} \in \Theta_{c_0}$ by Assumption \ref{ass: class of cmdp}.

For $k \geq C''$, we show that the Lyapunov function $\Phi(\lambda)=\lambda$ satisfies the conditions of Lemma \ref{lem:bandbound} and thus provide a bound on the exponential moment of the dual variable $\lambda$.

\begin{lemma}\label{lem:lyapcond}
For $k \geq C''$, when $\lambda \geq \varphi_k$, we have,
\begin{align*}
&\E \sbr{\lambda_{k+1} - \lambda_k | \lambda_k = \lambda} \leq \rho~\text{and} \\
&|\lambda_{k+1} -\lambda_k| \leq H \quad \text{with probability 1},
\end{align*}
where $\varphi_k := 4(H^2 + \epsilon_k^2 + \eta_k H)/(\tau - c^0)$ and \\ $\rho := - (\tau-c_0) / 4$. Thus, we have,
\begin{align}\label{eq:expmoment}
    \E\sbr{ e^{\zeta \lambda_{K+1}}} \le \E\sbr{e^{\zeta \lambda_{C''}}}+\frac{2 e^{\zeta\left(H+\varphi_{K+1}\right)}}{\zeta \rho},
\end{align}
where $\zeta=\rho/(H^{2}+H \rho / 3)$. The above inequality \eqref{eq:expmoment} can be simplified to
\begin{align}\label{eq:lbound}
   \E\sbr{\lambda_{K+1}} &\le \frac{1}{\zeta}\log\frac{11 H^2}{3 \rho^2} +  H + \sum_{1}^{C'' } \epsilon_k + C'' (H - \tau) \notag \\+ &\frac{4(H^2 + \epsilon_{K+1}^2 + \eta_{K+1} H)}{(\tau - c^0)}.
\end{align}
\end{lemma}

Next, we bound the $sum_k \epsilon_k$ term:
\begin{align}\label{eq:integral}
    &\sum_{k=1}^K \epsilon_k \geq \int_{1}^{K+1} \epsilon_u \mathrm{d} u \notag\\
    &\geq 10 H^{1.5} \sqrt{|\state|^2 |\A| K log |\state||\A|HK} \notag\\  
    &-10H^{1.5}\sqrt{|\state|^2 |\A| log |\state||\A|H}.
\end{align}

Thus, putting together \eqref{eq:firstsum}, \eqref{eq:lbound} and \eqref{eq:integral}, the leading terms of $\tilde{\mathcal{
O}}(\sqrt{K})$ cancel out and we get
$$\BR(K;c) = \tilde{\mathcal{
O}}\left(C''(H-\tau) + H^{1.5} \sqrt{|\state|^2 |\A| C''} \right) = \tilde{\mathcal{
O}}(1),$$
i.e., constraint violation regret is a constant, and does not grow with $K$.

\subsection{Reward Objective Regret Analysis}

We next provide regret analysis of the reward objective.
Let $\pi^{\epsilon_k,*}$ be the optimal policy for the pessimistic optimization problem (where $p$ is the true transition probability of the MDP):
    \begin{align}
   {\text{ max }} \quad & V_{1}^{\pi}(r,p)\\ \textrm{s.t.} \quad  & V_{1}^{\pi}(c,p) \leq \tau - \epsilon_k.\notag
\end{align}
Let $\pi^{\epsilon_k,\hat{p}_k}$ be the optimal policy for the pessimistic optimization problem (where $\hat{p}_k$ is the sampled transition probability of the MDP):
    \begin{align}
   {\text{ max }} \quad & V_{1}^{\pi}(r,\hat{p}_k)\\ \textrm{s.t.} \quad  & V_{1}^{\pi}(c,\hat{p}_k) \leq \tau - \epsilon_k.\notag
\end{align}
We can decompose the reward regret term as follows:
\begin{align}\label{eq: rewreg}
    &\BR(K;r) = \E\sbr{\sum_{k = 1}^K \left( V_{1}^{\pi^*}(r, p) - V^{\pi^k}_{1}(r, p) \right)}\notag\\
    &= \sum_{k = 1}^{C^{''}-1}
    \E\sbr{ V_{1}^{\pi^*}(r, p) - V^{\pi^k}_{1}(r, p)}
    \notag\\ 
    &+ \sum_{k = C''}^{K}
    \E\sbr{ V_{1}^{\pi^*}(r, p) - V^{\pi^k}_{1}(r, p)}\notag\\
    &\text{(splitting the sum across the sets of episodes)}\notag\\
    &= \sum_{k = 1}^{C^{''}-1}
    \E\sbr{ V_{1}^{\pi^*}(r, p) - V^{\pi_k}_{1}(r, p)}\notag\\ 
    &+ \sum_{k = C''}^{K}
    \E\sbr{ V_{1}^{\pi^*}(r, p) - V^{\pi^{\epsilon_k,*}}_{1}(r, p)} \notag\\
    &+ \sum_{k = C''}^{K}
    \E\sbr{ V_{1}^{\pi^{\epsilon_k,*}}(r, p) - V^{\pi^{\epsilon_k,\hat{p}_k}}_{1}(r, \hat{p}_k)} \notag\\
    &+\sum_{k = C''}^{K}
    \E\sbr{ V_{1}^{\pi^{\epsilon_k,\hat{p}_k}}(r, \hat{p}_k) - V^{\pi_k}_{1}(r, \hat{p}_k)} \notag\\
    &+\sum_{k = C''}^{K}
    \E\sbr{ V_{1}^{\pi_k}(r, \hat{p}_k) - V^{\pi_k}_{1}(r, p)}\notag\\
     &\text{(splitting the second sum into four parts)}\notag\\
    &\leq C^{''}H + \sum_{k = C''}^{K}
    \E\sbr{ V_{1}^{\pi^*}(r, p) - V^{\pi^{\epsilon_k,*}}_{1}(r, p)} \notag \\
    &+ \sum_{k = C''}^{K}
    \E\sbr{ V_{1}^{\pi^{\epsilon_k,*}}(r, p) - V^{\pi^{\epsilon_k,\hat{p}_k}}_{1}(r, \hat{p}_k)} \notag\\
    &+\sum_{k = C''}^{K}
    \E\sbr{ V_{1}^{\pi^{\epsilon_k,\hat{p}_k}}(r, \hat{p}_k) - V^{\pi_k}_{1}(r, \hat{p}_k)} \notag \\
    &+ \sum_{k = C''}^{K}
    \E\sbr{ V_{1}^{\pi_k}(r, \hat{p}_k) - V^{\pi_k}_{1}(r, p)}\notag\\
    &\leq C^{''}H + \sum_{k = C''}^{K}
    \E\sbr{ V_{1}^{\pi^*}(r, p) - V^{\pi^{\epsilon_k,*}}_{1}(r, p)} + 0 \notag\\
    &+\sum_{k = C''}^{K}
    \E\sbr{ V_{1}^{\pi^{\epsilon_k,\hat{p}_k}}(r, \hat{p}_k) - V^{\pi_k}_{1}(r, \hat{p}_k)} \notag \\ 
    &+  \sum_{k = C''}^{K}
    \E\sbr{ V_{1}^{\pi_k}(r, \hat{p}_k) - V^{\pi_k}_{1}(r, p)}\notag\\
    &(\text{by the posterior sampling property in Lemma \ref{lem: pseq}})\notag\\
    &\leq C^{''}H + \sum_{k = C''}^{K}
    \E\sbr{ V_{1}^{\pi^*}(r, p) - V^{\pi^{\epsilon_k,*}}_{1}(r, p)}\notag\\
    &+\sum_{k = C''}^{K}
    \E\sbr{ V_{1}^{\pi^{\epsilon_k,\hat{p}_k}}(r, \hat{p}_k) - V^{\pi_k}_{1}(r, \hat{p}_k)} \notag\\
    &+ H^{1.5}\sqrt{30|\state|^2|\A|K\log(|\state||\A|KH)} + 2H\notag\\
    &(\text{by the regret bound in Lemma \ref{lem: psrlreg}})\notag
\end{align}

The other terms are bounded as follows. Similar to Lemma 5.7 in \cite{liu2021learning}, we can define a  probabilistic mixed policy of $\pi^*$ and $\pi_{0}^{p}$ to prove the following lemma: 
\begin{lemma} \label{lem:kalathil}
The first summation term above can be bounded as
\begin{align}
  & \sum_{k=C''}^K \E\sbr{V_1^{\pi^*}(r, p) - V_1^{\pi^{\epsilon_k, *}}(r, p)} \notag \\
  &\le \sum_{k=C''}^K \frac{\epsilon_k H}{\tau - c^0} =\tilde{\mathcal{O}}\left(\frac{H^{2.5}}{\tau - c^0} \sqrt{|\state|^2 |\A| K}\right).  
\end{align}
\end{lemma}
By optimality of $\pi_k$ and the nature of the update of the dual parameter $\lambda_k$, we can prove the following lemma:
\begin{lemma}\label{lem:rew3}
    $$\sum_{k = C''}^{K}\E\sbr{ V_{1}^{\pi^{\epsilon_k,\hat{p}_k}}(r, \hat{p}_k) - V^{\pi_k}_{1}(r, \hat{p}_k)} = \tilde{\mathcal{
O}}\left(\frac{H}{\tau - c^0}\sqrt{K}\right)$$
\end{lemma}

The proof of this lemma can be found in the Appendix.

Now, putting together \eqref{eq: rewreg}, Lemma \ref{lem:kalathil} and lemma \ref{lem:rew3}, we get that
$$\BR(K;r) = \tilde{\mathcal{
O}}\left(\frac{H^{2.5}}{\tau - c^0} \sqrt{|\state|^2 |\A| K} \right).$$

\begin{remark}
We note that we can improve the upper bound on $\BR(K;r)$ from $\tilde{\mathcal{
O}}\left(H^{2.5}\sqrt{|\state|^2 |\A| K} \right)$ to $\tilde{\mathcal{
O}}\left( H^{2.5}\sqrt{|\state| |\A| K}\right)$ by using the leveraging an improved regret bound \cite{osband2017posterior} i.e., $\tilde{\mathcal{
O}}\left( H^{1.5}\sqrt{|\state||\A|K}\right)$ for the PSRL algorithm and appropriate scaling of the $\epsilon_k$ terms. But, this would require an assumption that the transition probability has an independent Dirichlet prior.
\end{remark}

\section{Experimental Results}\label{sec:expts}

In this section, we evaluate the empirical performance of the \texttt{Safe PSRL} algorithm and compare it with the state-of-the-art \texttt{DOPE} algorithm \cite{bura2022dope}, which 
% Note that the \texttt{DOPE} algorithm 
has been shown to perform better than other comparable algorithms (e.g., the \texttt{OptPess-LP} in \cite{liu2021learning}). 
The empirical performance is evaluated with respect to (i) the objective regret and (ii) the constraint regret. 

We consider the setting of a media streaming service \cite{bura2022dope} from a wireless base station. The base station provides the streaming service at two different speeds. These speeds follow independent Bernoulli distributions denoted by parameters $\mu_1 = 0.9$ and $\mu_2 = 0.1$, with $\mu_1$ corresponding to the faster service. The data packets arriving at the device are stored in a buffer and sent out according to a Bernoulli random process with mean $\gamma$. The buffer size $s_h$ evolves as $s_{h+1}=  \min \left( \max \left(0, s_h + A_h -B_h\right),N \right)$ where $A_h$ is the number of packet arrivals, $B_h$ is the number of packet departures, and $N =10$ is the maximum size of the buffer. The device desires to minimize the cost of running out of packets, i.e., an empty buffer, while restricting the use of the faster service. We model this scenario as a finite horizon CMDP with the state representing the buffer size and actions $\{1,2\}$ denoting the choice of speed. We set the objective cost as $r(s,a) = \mathbbm{1}\{s=0\}$ and the constraint cost as $c(s,a) = \mathbbm{1}\{a=1\}$. The episode length $H$ is 10 and the constraint threshold $\tau$ is 5.

We evaluate the cumulative regret for the \texttt{Safe PSRL} and the \texttt{DOPE} algorithm. The transition probability is fixed and not sampled from a prior distribution. For the \texttt{Safe PSRL} algorithm, we consider a  Dirichlet prior for the transition probability with parameters $[0.1,\ldots,0.1]$. The Dirichlet prior is a good choice since it is a conjugate prior for multinomial and categorical distributions. We further scale the $\epsilon_k$ parameters of the \texttt{Safe PSRL} algorithm by $0.05$ to avoid excessive pessimism.

The performance of our algorithm is compared against the \texttt{DOPE} algorithm, which requires a known safe policy. We choose the optimal policy of the given CMDP with a tighter constraint threshold $c_0 = 1$ as the safe policy. The same $c_0$ is also used in the 
\texttt{Safe PSRL} algorithm as the satisfiable constraint threshold.

The algorithms are evaluated over $K = 400,000$ episodes. All the experiments are performed on a 2019 MacBook Pro with 1.4 GHz Quad-Core Intel Core i5 processor and 16GB RAM.

\begin{figure}[h]
    \centering
    \includegraphics[width=0.475\linewidth]{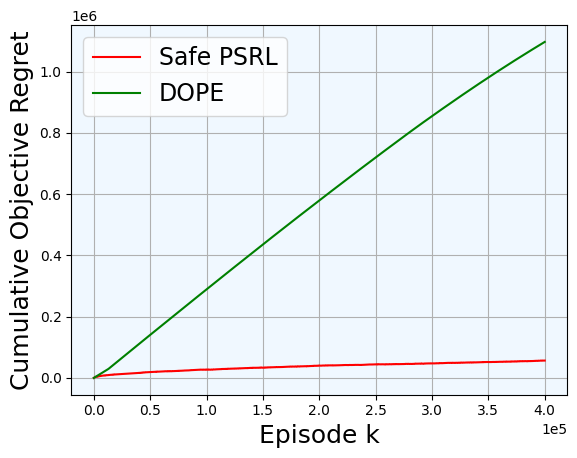}
    \hfill
    \centering
       \includegraphics[width=0.475\linewidth]{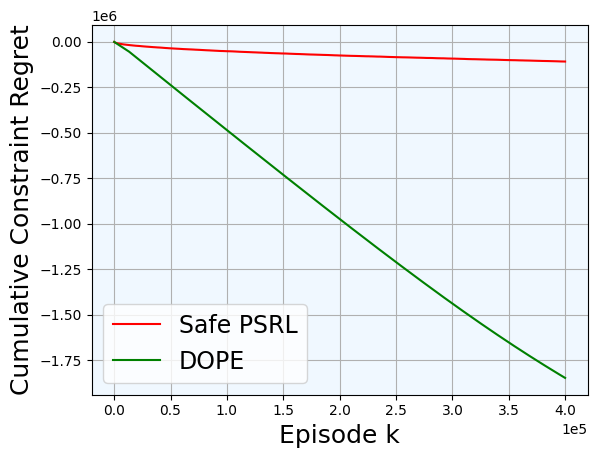}
    \caption{Plots showing (a) cumulative objective regret and (b) cumulative constraint regret for the \texttt{Safe PSRL} and \texttt{DOPE} algorithms.}
    \label{fig:reg}
\end{figure}

\begin{figure}[h]
    \centering
    \includegraphics[width=0.475\linewidth]{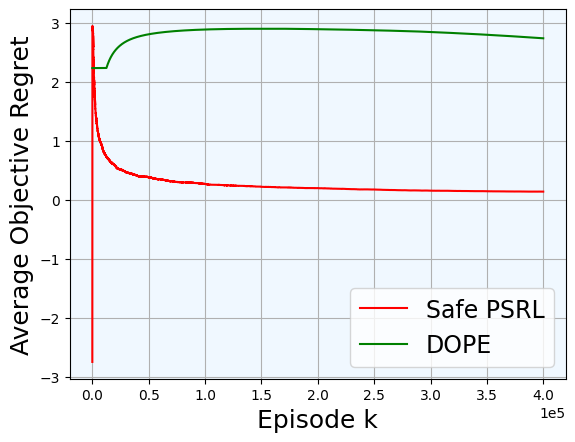}
    \hfill
    \centering
       \includegraphics[width=0.475\linewidth]{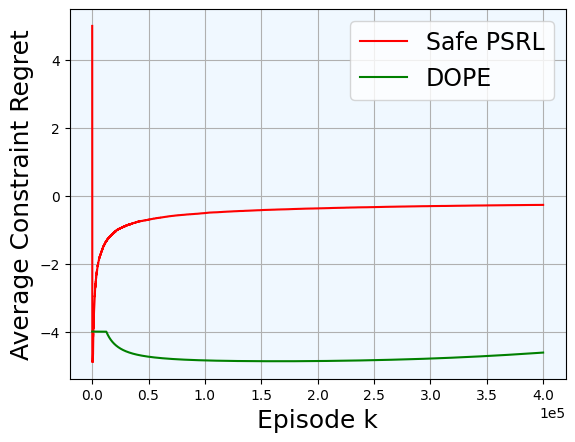}
    \caption{Plots showing (a) average objective regret and (b) average constraint regret for the \texttt{Safe PSRL} and \texttt{DOPE} algorithms.}
    \label{fig:avgreg}
\end{figure}

Fig.~\ref{fig:reg}(a) shows that the \texttt{Safe PSRL} algorithm greatly outperforms the \texttt{DOPE} algorithm in terms of objective regret. The objective regret for the \texttt{DOPE} algorithm grows almost linearly for a very large number of episodes. In comparison, the  \texttt{Safe PSRL} attains $\sqrt{K}$ behavior much earlier. Fig.~\ref{fig:avgreg}(a) for the average objective regret shows this behavior more clearly.

In Fig.~\ref{fig:reg}(b), we see that the constraint regrets for both the \texttt{Safe PSRL} and the \texttt{DOPE} algorithm are negative for almost all of the episodes. This implies that the constraint was satisfied in almost all of the episodes and matches with the theoretical guarantees for both the algorithms.  

We observe the initial jumps in Fig.~\ref{fig:avgreg}(a) and Fig.~\ref{fig:avgreg}(b) with respect to the \texttt{Safe PSRL} regret plots because a few initial policies returned by the \texttt{Safe PSRL} algorithm fail to satisfy the constraint while achieving better reward objective performance. 

This behavior occurs because the dual parameter $\lambda$, which starts from $0$, has not yet caught up with the appropriate value which would ensure optimal objective performance while satisfying the constraint. We can infer from the regret plot that this appropriate $\lambda$ value is reached fairly quickly by the  \texttt{Safe PSRL} algorithm. The \texttt{DOPE} algorithm, on the other hand, relies on the safe policy for too long before it starts to explore.

We thus show that the \texttt{Safe PSRL} algorithm is able to achieve superior objective regret performance while satisfying the constraint for almost all the episodes. This result is further achieved without the knowledge of a safe policy.

\section{Conclusions}\label{sec:conclusions}

We addressed the problem of safe online learning for episodic MDPs with constraints and unknown transition probabilities. The \texttt{Safe PSRL} is the first posterior sampling algorithm that achieves bounded constraint violation regret while achieving near-optimal cumulative reward regret. The algorithm has better empirical performance than other state-of-the-art algorithms (e.g., the \texttt{DOPE} algorithm) for the same setting and does not need to assume knowledge of a safe policy. The algorithm can be extended to the infinite-horizon setting. Incorporating chance or risk constraints would be another interesting direction for future work.

\newpage
\bibliography{ref}
\bibliographystyle{icml2023}
\newpage
\appendix
%\onecolumn
\section{Proofs}
\subsection{Proof of Lemma \ref{lem:lyapcond}}
\begin{proof}
Now for $k \geq C''$, consider:
\begin{align*}
    &\frac{{\lambda_{k+1}}^2}{2} - \frac{{\lambda_k}^2}{2}\\ 
    &= \lambda_k(\lambda_{k+1} - \lambda_k) + \frac{1}{2}(\lambda_{k+1} - \lambda_k)^2\\
    &= \lambda_k (V^{\pi_k}_{1}( c,\hat{p}_k) + \epsilon_k- \tau) + \frac{1}{2}(V^{\pi_k}_{1}( c,\hat{p}_k) + \epsilon_k- \tau)^2\\
    &= \lambda_k (V^{\pi_k}_{1}( c,\hat{p}_k) + \epsilon_k- \tau) -\eta_k V^{\pi_k}_{1}( r,\hat{p}_k) \\
    &+ \eta_k V^{\pi_k}_{1}( r,\hat{p}_k) + \frac{1}{2}(V^{\pi_k}_{1}( c,\hat{p}_k) + \epsilon_k- \tau)^2\\
    &\leq \lambda_k (V^{\pi_k}_{1}( c,\hat{p}_k) + \epsilon_k- \tau) -\eta_k V^{\pi_k}_{1}( r,\hat{p}_k) \\
    &+ \eta_k H + \frac{1}{2}(V^{\pi_k}_{1}( c,\hat{p}_k) + \epsilon_k- \tau)^2\\
    &\leq \lambda_k (V^{\pi_k}_{1}( c,\hat{p}_k) + \epsilon_k- \tau) -\eta_k V^{\pi_k}_{1}( r,\hat{p}_k) \\
    &+ \eta_k H + (V^{\pi_k}_{1}( c,\hat{p}_k) - \tau)^2 + \epsilon_{k}^{2} \\
    &(\text{ Using } \frac{(a+b)^2}{2} \leq a^2 + b^2 )\\
    &\leq \lambda_k (V^{\pi_k}_{1}( c,\hat{p}_k) + \epsilon_k- \tau) -\eta_k V^{\pi_k}_{1}( r,\hat{p}_k) \\
    &+ \eta_k H + H^2 + \epsilon_{k}^{2} \\
    &\leq  \lambda_k (V^{\pi_{0}^{\hat{p}_k}}_{1}( c,\hat{p}_k) + \epsilon_k- \tau) -\eta_k V^{\pi_{0}^{\hat{p}_k}}_{1}( r,\hat{p}_k) \\
    &+ \eta_k H + H^2 + \epsilon_{k}^{2} \\
    &(\text{ By optimality of $\pi_k$ in primal update })\\
    &\leq \lambda_k (c_0 + \epsilon_k - \tau ) + \eta_k H + H^2 + \epsilon
_{k}^{2}\\
&\leq -\frac{\lambda_k(\tau-c_0)}{2}+ \eta_k H + H^2 + \epsilon
_{k}^{2}  \\
&(\text{ as for } k\geq C'', \epsilon_k \leq \frac{(\tau-c
_0)}{2} ) 
\end{align*}

Now for $\lambda \geq \varphi_k$ where $\varphi_k := 4(H^2 + \epsilon_k^2 + \eta_k H)/(\tau - c^0)$, we have:
\begin{align*}
    &\E \sbr{\lambda_{k+1} - \lambda_k | \lambda_k = \lambda} \leq \E \sbr{\frac{\lambda_{k+1}^{2} - \lambda_{k}^{2}}{2 \lambda_k}| \lambda_k = \lambda }\\
    &(\text{Using } x - y \leq \frac{x^2-y^2}{2y},\text{for } y > 0)\\
    &= \frac{1}{\lambda} \E \sbr{\frac{\lambda_{k+1}^{2} - \lambda_{k}^{2}}{2 }| \lambda_k = \lambda }\\
    &\leq\frac{1}{\lambda}\E\sbr{-\frac{\lambda_k(\tau-c_0)}{2}+ \eta_k H + H^2 + \epsilon
_{k}^{2} | \lambda_k = \lambda}\\
&= -\frac{(\tau-c_0)}{2}+\frac{\eta_k H + H^2 + \epsilon
_{k}^{2}}{\lambda}\\
&\leq -\frac{(\tau-c_0)}{2} +\frac{(\tau-c_0)}{4}\\
&=-\frac{(\tau-c_0)}{4} := \rho
\end{align*}

Further, $|\lambda_{k+1} -\lambda_k| = |V^{\pi_k}_{1}( c,\hat{p}_k) + \epsilon
_k - \tau| \leq H $ with probability 1. Thus, by lemma \ref{lem:bandbound}, we have :
\begin{align*}
    \E\sbr{ e^{\zeta \lambda_{K+1}}} \le \E\sbr{e^{\zeta \lambda_{C''}}}+\frac{2 e^{\zeta\left(H+\varphi_{K+1}\right)}}{\zeta \rho},
    \end{align*}
    where $\zeta=\rho/(H^{2}+H \rho / 3)$.
\begin{align*}
    &\implies e^{\zeta \E\sbr{\lambda_{K+1}}} \leq \E\sbr{e^{\zeta \lambda_{C''}}}+\frac{2 e^{\zeta\left(H+\varphi_{K+1}\right)}}{\zeta \rho} \\
    &\text{ (By Jensen's inequality) }\\
    &\implies \E\sbr{\lambda_{K+1}} \leq \frac{1}{\zeta}\log \sbr{\E\sbr{e^{\zeta \lambda_{C''}}}+\frac{2 e^{\zeta\left(H+\varphi_{K+1}\right)}}{\zeta \rho}}\\
\end{align*}
Further, 
\begin{align*}
\lambda_{C''} &\leq \lambda_1 + \sum_{1}^{C'' - 1} (V^{\pi^k}_{1}(c, \hat{p}_k) + \epsilon_k - \tau)_{+} \\&\leq  \sum_{1}^{C'' } \epsilon_k + C'' (H - \tau)\\
&:= \lambda_{C''}^{\max}
\end{align*}
Continuing, 
\begin{align}
    &\E\sbr{\lambda_{K+1}} \notag
    \\&\leq \frac{1}{\zeta}\log \sbr{e^{\zeta \lambda_{C''}^{\max}}+\frac{2 e^{\zeta\left(H+\varphi_{K+1}\right)}}{\zeta \rho}}\notag\\
    &\leq \frac{1}{\zeta}\log \sbr{e^{\zeta \lambda_{C''}^{\max}}+\frac{8H^2 e^{\zeta\left(H+\varphi_{K+1}\right)}}{3 \rho^2}}\notag\\
    &(\text{ Using } \zeta \geq \frac{3(\tau-c_0)}{13 H^2})\notag\\
    &\leq \frac{1}{\zeta}\log \sbr{\frac{11 H^2}{3 \rho^2}e^{\zeta\left(H+\varphi_{K+1} + \lambda_{C''}^{\max} \right)}}\notag\\
    &= \frac{1}{\zeta}\log\frac{11 H^2}{3 \rho^2} +  H+\varphi_{K+1} + \lambda_{C''}^{\max} \notag\\
    &= \frac{1}{\zeta}\log\frac{11 H^2}{3 \rho^2} +  H + \sum_{1}^{C'' } \epsilon_k + C'' (H - \tau) \notag \\
    &+ \frac{4(H^2 + \epsilon_{K+1}^2 + \eta_{K+1} H)}{(\tau - c^0)}\notag
\end{align}
\end{proof}
\subsection{Proof of Lemma \ref{lem:rew3}}
\begin{proof}
   \begin{align*}
       &\sum_{k = C''}^{K}\E\sbr{ V_{1}^{\pi^{\epsilon_k,\hat{p}_k}}(r, \hat{p}_k) - V^{\pi_k}_{1}(r, \hat{p}_k)} \\
       &= \sum_{k = C''}^{K} \E \sbr{\frac{\lambda_k}{\eta_k}\left( V_{1}^{\pi^{\epsilon_k,\hat{p}_k}}(c, \hat{p}_k) - V^{\pi_k}_{1}(c, \hat{p}_k) \right)} \\ 
       &+\sum_{k = C''}^{K} \E \left[\left(V_{1}^{\pi^{\epsilon_k,\hat{p}_k}}(r, \hat{p}_k) -  \frac{\lambda_k}{\eta_k}V_{1}^{\pi^{\epsilon_k,\hat{p}_k}}(c, \hat{p}_k)\right) \right]\\
       &-\sum_{k = C''}^{K} \E\left[ \left( V_{1}^{\pi^{k}}(r, \hat{p}_k) -  \frac{\lambda_k}{\eta_k}V_{1}^{\pi^{k}}(c, \hat{p}_k)\right) \right]\\
       &\leq \sum_{k = C''}^{K} \E \sbr{\frac{\lambda_k}{\eta_k}\left( V_{1}^{\pi^{\epsilon_k,\hat{p}_k}}(c, \hat{p}_k) - V^{\pi_k}_{1}(c, \hat{p}_k) \right)} + 0\\
       &(\text{ By optimality of $\pi_k$ in primal update })\\
       &\leq \sum_{k = C''}^{K} \E \sbr{\frac{\lambda_k}{\eta_k}\left( \tau - \epsilon_k -  V^{\pi_k}_{1}(c, \hat{p}_k) \right)}\\
       &\leq \sum_{k = C''}^{K} \E\sbr{\frac{1}{\eta
_k}((\lambda_k(\lambda_{k+1} - \lambda_k) + \tau^2)}\\
&\text{(By update rule for $\lambda_k$)}\\
&\leq \E \left[ \sum_{k = C''}^{K} \frac{1}{\eta_k}(\frac{\lambda_k^2}{2} - \frac{\lambda_{k+1}^2}{2})+\sum_{k = C''}^{K} \frac{1}{2\eta_k}(\lambda_{k+1} - \lambda_k)^2 \right.\\
&\left.+ \sum_{k = C''}^{K}\frac{\tau^2}{\eta_k} \right]\\
&\leq \E \sbr{\frac{(\lambda_{C''})^2}{2\eta_{C''}}} + \sum_{k = C''}^{K} \frac{H^2}{2\eta_k}+ \sum_{k = C''}^{K}\frac{H^2}{\eta_k} \\
&\text{ (As $\eta_k$ increases with $k$)}\\
&\leq \frac{(\sum_{k = 1}^{C''}\epsilon_k + C^{''}(H-\tau))^2}{2\eta_{C''}} + \frac{3H}{2}\sum_{C''}^{K}\frac{1}{(\tau - c_0)\sqrt{k}}\\
&= \tilde{\mathcal{
O}}\left(\frac{H}{\tau - c^0}\sqrt{K}\right)
   \end{align*}
\end{proof}

\end{document}